\documentclass[letterpaper]{article} % DO NOT CHANGE THIS
\usepackage{aaai24}  % DO NOT CHANGE THIS
\usepackage{times}  % DO NOT CHANGE THIS
\usepackage{helvet}  % DO NOT CHANGE THIS
\usepackage{courier}  % DO NOT CHANGE THIS
\usepackage[hyphens]{url}  % DO NOT CHANGE THIS
\usepackage{graphicx} % DO NOT CHANGE THIS
\usepackage{natbib}  % DO NOT CHANGE THIS AND DO NOT ADD ANY OPTIONS TO IT
\usepackage{caption} % DO NOT CHANGE THIS AND DO NOT ADD ANY OPTIONS TO IT
\usepackage{algorithm}
\usepackage{algorithmic}

\usepackage{newfloat}
\usepackage{listings}
\DeclareCaptionStyle{ruled}{labelfont=normalfont,labelsep=colon,strut=off} % DO NOT CHANGE THIS
\floatstyle{ruled}
\newfloat{listing}{tb}{lst}{}
\floatname{listing}{Listing}
\usepackage{booktabs}
\usepackage{tabularx}
\usepackage{multirow}
\usepackage{amsthm}
\usepackage{amsmath}
\usepackage{amssymb}
\usepackage{utfsym}

\title{VMT-Adapter: Parameter-Efficient Transfer Learning for Multi-Task Dense Scene Understanding}
\author{
    Yi Xin\textsuperscript{\rm 1,2}\equalcontrib,
    Junlong Du\textsuperscript{\rm2}\equalcontrib,
    Qiang Wang\textsuperscript{\rm2},
    Zhiwen Lin\textsuperscript{\rm2},
    Ke Yan\textsuperscript{\rm2}\thanks{Corresponding author.}
}
\affiliations{
    \textsuperscript{\rm 1}State Key Laboratory for Novel Software Technology, Nanjing University, Nanjing, China\\
    \textsuperscript{\rm 2}Youtu Lab, Tencent\\
    xinyi@smail.nju.edu.cn, jeffdu@tencent.com, albertqwang@tencent.com, \\xavierzwlin@tencent.com, kerwinyan@tencent.com
}

\usepackage{bibentry}
\begin{document}

\maketitle

\begin{abstract}
Large-scale pre-trained models have achieved remarkable success in various computer vision tasks. A standard approach to leverage these models is to fine-tune all model parameters for downstream tasks, which poses challenges in terms of computational and storage costs. Recently, inspired by Natural Language Processing (NLP), parameter-efficient transfer learning has been successfully applied to vision tasks. However, most existing
techniques primarily focus on single-task adaptation, and despite limited
research on multi-task adaptation, these methods often exhibit suboptimal
training and inference efficiency. In this paper, we first propose an once-for-all \textit{Vision Multi-Task Adapter (VMT-Adapter)}, which strikes approximately $O(1)$ training and inference efficiency w.r.t task number. Concretely, \textit{VMT-Adapter} shares the knowledge from multiple tasks to enhance cross-task interaction while preserves task-specific knowledge via independent knowledge extraction modules. Notably, since task-specific modules require few parameters, \textit{VMT-Adapter} can handle an arbitrary number of tasks with a negligible increase of trainable parameters. We also propose \textit{VMT-Adapter-Lite}, which further reduces the trainable parameters by learning shared parameters between down- and up-projections. Extensive experiments on four dense scene understanding tasks demonstrate the superiority of \textit{VMT-Adapter(-Lite)}, achieving a 3.96\% 
 (1.34\%) relative improvement compared to single-task full fine-tuning, while utilizing merely $\sim1\%$ ($0.36\%$) trainable parameters of the pre-trained model.
\end{abstract}

\section{Introduction}
\label{sec:intro}
The pretrain-finetune paradigm has made significant strides in Natural Language Processing~(NLP)~\cite{devlin2018bert, brown2020language}, Computer Vision~(CV)~\cite{he2022masked, xie2022simmim}, and various other domains. Typically, given a pre-trained model, the conventional fine-tuning approach involves adjusting the entire model, that is, performing full fine-tuning for downstream tasks. However, as state-of-the-art pre-trained models expand to encompass billions or even trillions of parameters, the traditional method of full fine-tuning becomes increasingly untenable due to the immense computational and storage resource demands. 

\begin{figure}[t]
    \includegraphics[width=\linewidth]{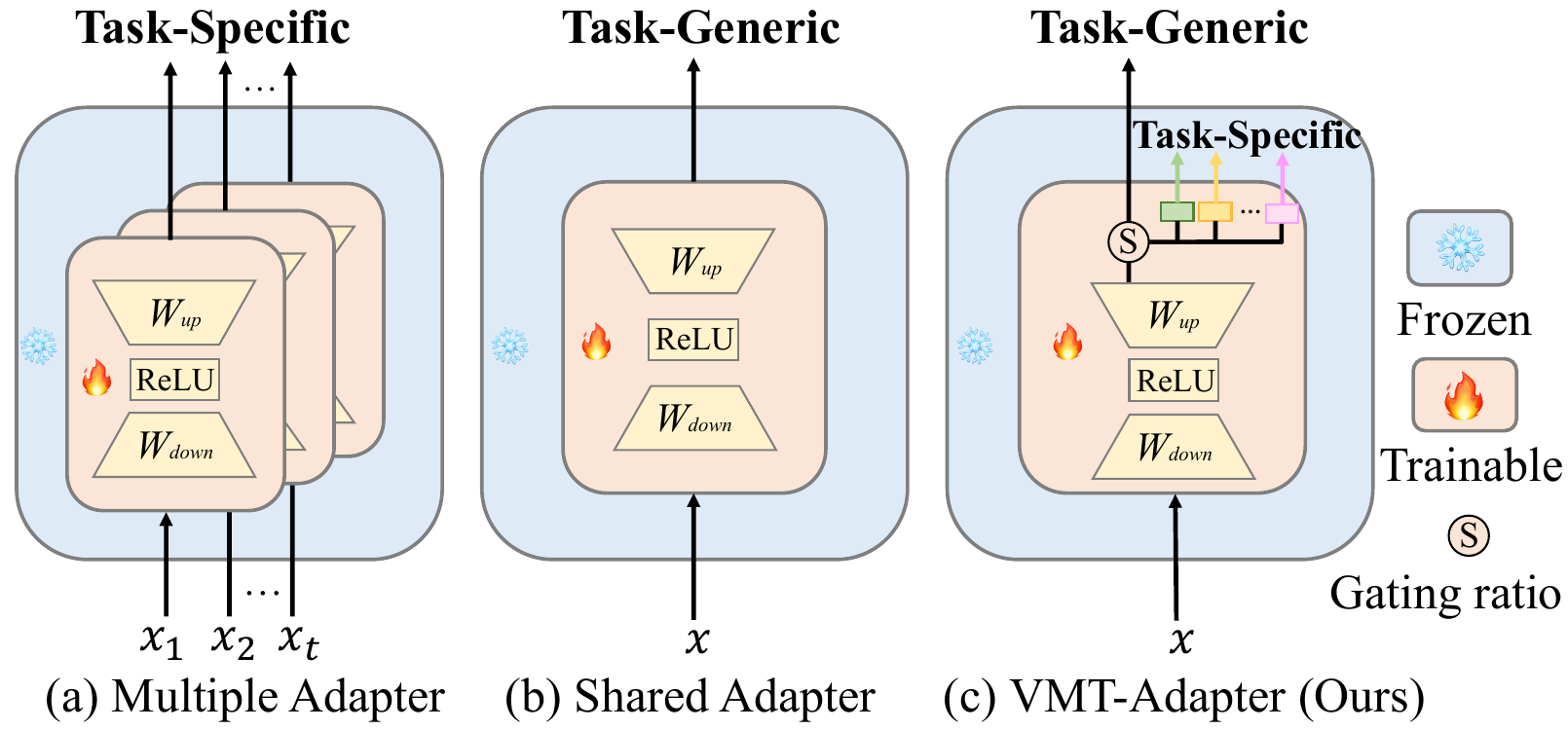}
    \caption{
    Illustration of (a) Multiple Adapter: inject separate adapters for each task and (b) Shared Adapter: $t$ tasks share one adapter. (c) VMT-Adapter (Ours): $t$ tasks share the down- and up-projections to enhance cross-task interaction and split the cross-task representations into task-generic and the input of task-specific knowledge extraction modules.
    }
    \label{fig:adapters}
\end{figure}

To address these challenges, researchers have extensively investigated parameter-efficient transfer learning approaches~\cite{houlsby2019parameter, hu2021lora, zaken2021bitfit}, which strive to achieve an optimal balance between trainable parameters and performance on downstream tasks. Among these methods, the Adapter~\cite{houlsby2019parameter} and its variants~\cite{karimi2021parameterefficient, mahabadi2021compacter} have gained widespread adoption in the NLP domain and have been integrated into various architectures. The Adapter is a compact module incorporated into the intermediate layers of the model (as depicted in Figure~\ref{fig:adapters}), enabling comparable performance to full fine-tuning while only training a limited set of parameters. This innovative approach significantly reduces the computational and storage overhead, making it a more practical solution for large-scale models and diverse applications.
\begin{figure}[t]
   \begin{picture}(0,196)
     \put(-14,-10){\includegraphics[width=1.1\linewidth]{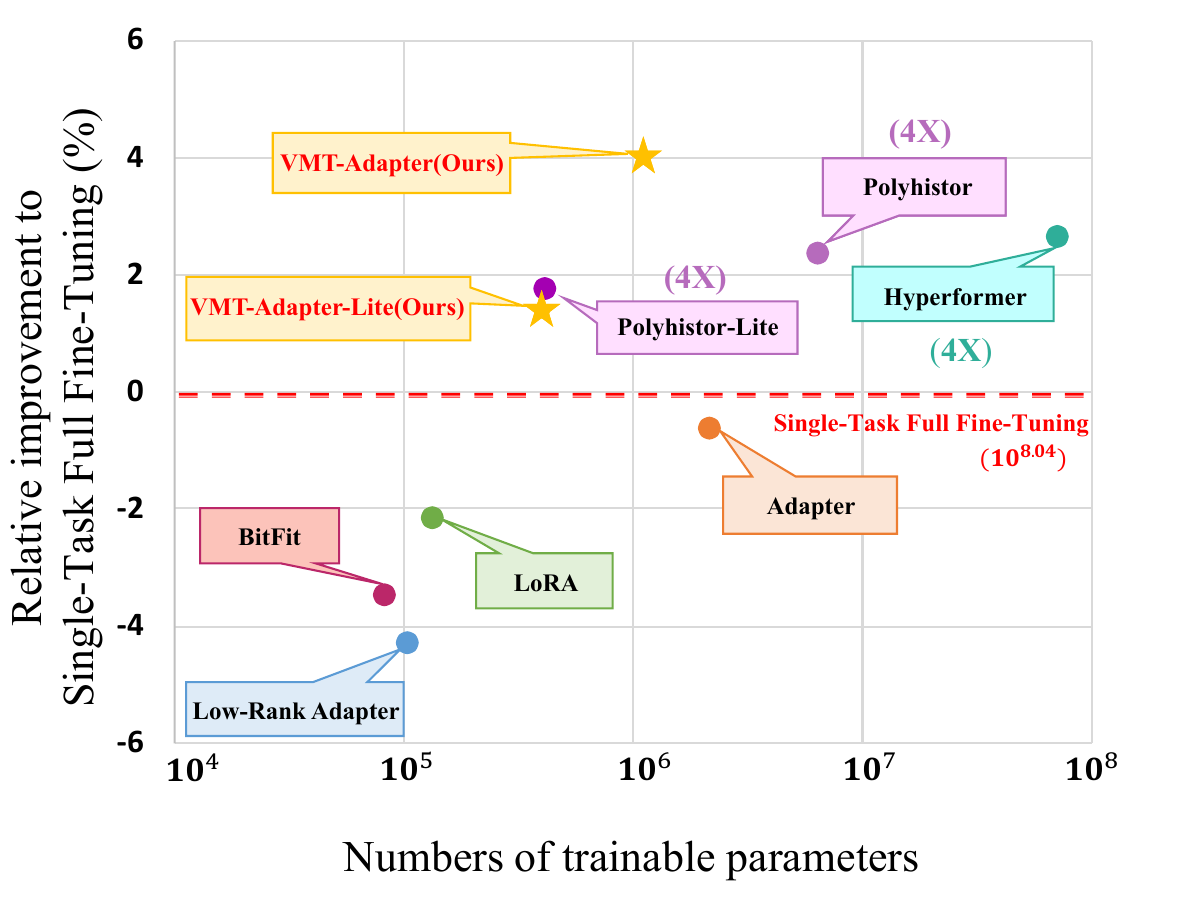}}
   \end{picture}
    \caption{The trade-off between performance and trainable parameters of various parameter-efficient tuning methods. The result is the average performance on four dense tasks. $4X$ means that training and inference costs is four times.}
    \label{fig:pipeline}
\end{figure}
Recently, Adaptformer~\cite{chen2022adaptformer} introduced adapter into computer vision, focusing on visual recognition tasks. Owing to its success, parameter-efficient transfer learning for vision has garnered considerable interest. While some subsequent studies~\cite{he2022parameter,yu2022towards} have demonstrated promising results, these approaches primarily concentrate on single-task adaptation. For multiple downstream tasks, the individual training and storage of task-specific parameters prove to be inefficient, as the trainable parameters grows in proportion to the number of tasks. Consequently, further exploration is warranted to determine how pre-trained models can be transferred in a parameter-efficient manner to tackle more challenging, simultaneous multi-task vision problems, such as semantic segmentation, surface normal estimation, and saliency detection.

In multi-task parameter-efficient transfer learning, one intuitive approach is to incorporate adapters (or other single-task adaptation methods) for each task, referred to as \textit{Multiple Adapter}, as illustrated in Figure~\ref{fig:adapters}a. However, this method results in a linear increase in the number of adapters with respect to the task number, leading to relatively high parameters. Moreover, the lack of task interactions among these adapters may lead to suboptimal performance. An alternative approach is to employ a shared adapter (or other single-task adaptation methods) across tasks, known as \textit{Shared Adapter}, as illustrated in Figure~\ref{fig:adapters}b. While this method reduces trainable parameters and promotes interaction between tasks, it may lack task-specific knowledge for each individual task. Polyhistor~\cite{liu2022polyhistor} is the first to address parameter-efficient multi-task transfer learning, building upon the \textit{Multiple Adapter}. It reduces trainable parameters by sharing adapter parameters across tasks and layers. However, there are still some limitations. Due to the design of task-independent adapters, the number of adapters, as well as training and inference costs, grow linearly with task number. Furthermore, Polyhistor involves numerous hyperparameters, which may hinder rapid application and deployment.

To overcome the aforementioned limitations, we propose a novel parameter-efficient \textit{VMT-Adapter}, which comprises shared projections and knowledge extraction modules for adapting to multiple vision tasks. The \textit{VMT-Adapter} unifies the learning of task-generic and task-specific representations within a single framework. Notably, the shared projections parameters are independent of task number, while the knowledge extraction modules merely contains extremely few parameters. As a result, \textit{VMT-Adapter} is well-suited for multi-task transfer learning and can be seamlessly integrated into any transformer intermediate layer.
Furthermore, drawing inspiration from Low-Rank Adapter~\cite{sung2022vladapter} and Compactor~\cite{mahabadi2021compacter}, we develop the \textit{VMT-Adapter-Lite}, which reduces the parameters to $\frac{1}{m}$ of the original (where $m$ represents the dimension of the shared matrices, as detailed in Section~\ref{section:VMT-Adapter-Lite}). This is achieved by sharing learnable parameters between the down- and up-projections within the \textit{VMT-Adapter}.

To establish a multi-task parameter-efficient transfer learning benchmark, Liu et al.~\cite{liu2022polyhistor} adapt various single-task adaptation methods from NLP to vision multi-task settings. Building upon their work, we extend these baselines and compare them with state-of-the-art approaches. Our experiments are conducted on the PASCAL-Context dataset, focusing on dense scene understanding tasks. As depicted in Figure~\ref{fig:pipeline}, the results highlight the effectiveness of \textit{VMT-Adapter(-Lite)}, striking an optimal balance between performance and trainable parameters. 

Our main contributions are as follows:
\begin{itemize}
    \item We propose \textit{VMT-Adapter}, a novel parameter-efficient adapter for multi-task dense scene understanding, which learns task-generic and task-specific representations in an unified framework. To the best of our knowledge, \textit{VMT-Adapter} is the first once-for-all adapter for vision multi-task learning with approximately $O(1)$ training and inference efficiency w.r.t task number.
    \item We propose a light-weight version termed \textit{VMT-Adapter-Lite} via a parameter-sharing strategy to meet more efficient requirements.
    \item Experimental results on dense scene understanding tasks show that \textit{VMT-Adapter(-Lite)} achieves competitive performance compared to the single-task full fine-tuning leveraging merely $\sim1\%$ ($0.36\%$) of the pre-trained model parameters, as shown in Figure~\ref{fig:pipeline}.
\end{itemize}

\section{Related Work}
\label{sec:relate}
\paragraph{Transformer in Vision.}
Transformer was initially introduced for Natural Language Processing (NLP) tasks, such as machine translation~\cite{vaswani2017attention} and text generation~\cite{devlin2018bert}. Its remarkable success in these domains has inspired a shift in computer vision research towards Transformer-based models, beginning with the introduction of the Vision Transformer (ViT) ~\cite{dosovitskiy2020image}. Since then, a variety of Transformer-based models~\cite{liu2021swin, xie2021segformer} have demonstrated impressive performance across a wide range of vision tasks, including image classification, semantic segmentation, object detection, etc. To further enhance the performance of downstream tasks and reduce the consumption of training resources, researchers have provided pre-trained ViT-based models on large-scale datasets, such as ImageNet~\cite{russakovsky2015imagenet}. By fine-tuning these pre-trained models on downstream tasks, researchers have consistently achieved faster convergence\cite{he2019rethinking} and improved performance, showcasing the potential of Transformer-based models in computer vision.

\paragraph{Multi-Task Dense Scene Understanding.} 
Multi-Task Learning (MTL) aims to simultaneously learn multiple tasks by sharing knowledge and computation. Numerous studies~\cite{li2022learning, xu2023demt} have demonstrated that various dense vision tasks, which are more challenging than image classification, can benefit from multi-task learning. Current research on multi-task dense scene understanding primarily focuses on model architecture design, which can be categorized into encoder-based and decoder-based methods.
Encoder-based methods~\cite{gao2019nddr, liu2019end} are dedicated to designing task interaction modules embedded within the encoder, while decoder-based methods~\cite{bruggemann2021exploring, zhang2021transfer} focus on module design at the decoder stage. As vision pre-trained models become increasingly powerful, encoders directly adopt models such as the Vision Transformer (ViT)~\cite{dosovitskiy2020image} and Swin Transformer~\cite{liu2021swin}, which are pre-trained on large-scale image datasets. This approach leads to significant performance improvements and has become the mainstream for multi-task.
However, the pretrain-finetune paradigm has two notable drawbacks. First, fine-tuning the encoder results in an unavoidable computational cost, as all model parameters must be adjusted. Second, a shared encoder across multiple tasks primarily extracts task-generic knowledge, while task-specific knowledge is neglected.

\paragraph{Parameter-Efficient Transfer Learning.}
Parameter Efficient Transfer Learning aims to adapt a pre-trained model to downstream tasks by training only a small number of parameters. The most straightforward approach involves freezing the pre-trained encoder and fine-tuning the last layer. However, this method often falls short of full fine-tuning in terms of downstream task accuracy. 
Several studies~\cite{zaken2021bitfit} have attempted to improve linear probing performance by updating the bias term in all layers. In contrast, other works~\cite{chen2022adaptformer, he2022towards} have proposed to insert adapters into transformer layers for fine-tuning in a parameter-efficient manner. More recently, LoRA~\cite{hu2021lora} introduced a method that generates two low-rank matrices, which are then multiplied and serve as a residual of attention weight matrices.
While these methods have demonstrated satisfactory performance with few trainable parameters, they primarily focus on single-task transfer learning. For multi-task parameter-efficient transfer learning, Polyhistor~\cite{liu2022polyhistor} incorporated task-independent adapters into each transformer layer and further reduced trainable parameters through a parameter-sharing strategy between adapters across different layers and tasks. However, the multi-task adapter structure design itself has not been thoroughly explored, leading to separate adapters for each task and increased training and inference costs. Consequently, there is a linear relationship between the number of adapters and tasks.

% Another study~\cite{sung2022vladapter} shares a single adapter across multiple tasks, which can negatively impact task performance when conflicts arise between tasks.

\section{Background}

% \paragraph{Vision Transformer.} Fine-tuning large-scale pretrained Vision Transformer (ViT)~\cite{dosovitskiy2020image} has demonstrated superior performance on computer vision tasks. The hierarchical vision transformer, a variant of ViT, is more effective on dense scene understanding tasks. In this work, we adopt SwinTransformer~\cite{liu2021swin}, which consists of four hierarchical blocks. Each block has several transformer layers, and each transformer layer comprises a Shifted Window-based Multi-head Self-Attention (SW-MSA) module and a fully connected Feed-Forward Network (FFN) implemented by a 2-layer MLP. Layer Normalization (LN) and residual connection are performed before and after the FFN and SW-MSA modules, as shown in Figure~\ref{fig:background}.

\begin{figure*}[t]
    \centering
    \includegraphics[width=0.9\linewidth]{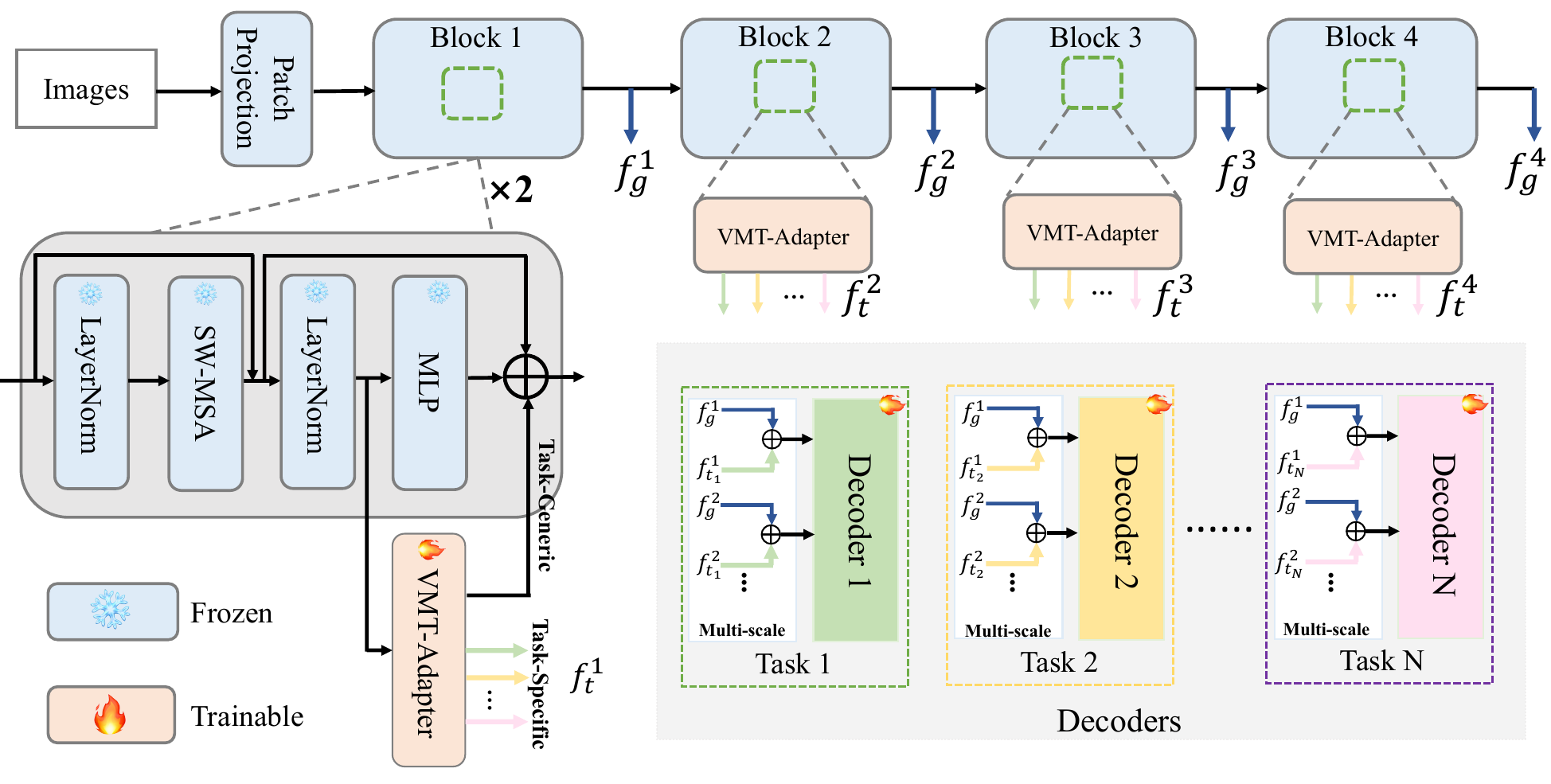}
    \caption{
    Illustration of SwinTransformer architecture and how to use VMT-Adapter. We insert VMT-Adapter in parallel to the MLP layer. The decoder of each task receives multi-scale information from different transformer blocks, including task-generic and task-specific representations. \includegraphics[width=0.32cm]{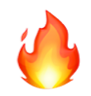} represents trainable parameters, \includegraphics[width=0.32cm]{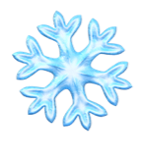} represents frozen parameters.
    }
    \label{fig:background}
\end{figure*}

\paragraph{Hierarchical Vision Transformer.} The hierarchical vision transformer, a variant of ViT~\cite{dosovitskiy2020image}, is more effective on dense scene understanding tasks. In this work, we adopt SwinTransformer~\cite{liu2021swin}, which consists of four hierarchical blocks. Each block has several transformer layers, and each transformer layer comprises a Shifted Window-based Multi-head Self-Attention (SW-MSA) module and a fully connected Feed-Forward Network (FFN) implemented by a 2-layer MLP. Layer Normalization (LN) and residual connection are performed before and after the FFN and SW-MSA modules, as shown in Figure~\ref{fig:background}.

\paragraph{Adapter.} Adapter~\cite{houlsby2019parameter} is a bottleneck-like architecture that consists of a down-projection layer $W_{down}\in R^{d\times k}$ and an up-projection layer $W_{up}\in R^{k\times d}$, where $k$ reduces the dimension of representation $d$ into low-rank $(k<<d)$. Additionally, there is a ReLU layer between the two layers for non-linear projection. In general, Adapter is injected into transformer layers and updated during training, while the parameters of the transformer are frozen. Given a specific input feature $x_{\ell} \in R^{d}$, Adapter generates the adapted features with a residual connection:
\begin{equation}
% \begin{align}
    \tilde{x}_{\ell}=\operatorname{ReLU}\left(x_{\ell} \cdot W_{down}\right) \cdot W_{up}+x_{\ell},
% \end{align}
\end{equation}
where $\tilde{x}_{\ell}$ is the output, and $W=[W_{down};W^{T}_{up}] \in R^{d \times 2k}$ denotes all trainable parameters.

\paragraph{Kronecker Product.}
The Kronecker Product between matrix $\mathbf{A}\in R^{a_1\times a_2}$ and $\mathbf{B}\in R^{b_1\times b_2}$ yields a block matrix $W\in R^{w_1\times w_2}$, where $w_1=a_1 \times b_1$ and $w_2=a_2 \times b_2$. In $W$, each block $(i,j)$ is the result of multiplying the element $a_{ij}$ with matrix $\mathbf{B}$, which is defined as:
\begin{equation}
\mathbf{W}=\mathbf{A} \otimes \mathbf{B}=\left(\begin{array}{ccc}
a_{11} \mathbf{B} & \cdots & a_{1 n} \mathbf{B} \\
\vdots & \ddots & \vdots \\
a_{m 1} \mathbf{B} & \cdots & a_{m n} \mathbf{B}
\end{array}\right).    
\end{equation}

\section{Method}
\label{sec:method}
\paragraph{Overall.} Our goal is to efficiently adapt a large-scale pre-trained model for multi-task scenarios by introducing only a few additional trainable parameters. We enhance the multi-task parameter-efficient transfer learning process in two significant ways: (1) We design VMT-Adapter tailored for multi-task characteristics, considering both task-generic and task-specific representations while dramatically reducing the number of trainable parameters. To the best of our knowledge, this is the first once-for-all adapter structure designed for multi-task dense scene understanding. (2) A parameter-sharing method between down-projection and up-projection inside VMT-Adapters is proposed to further reduce the trainable parameters. 

\subsection{VMT-Adapter}
\label{section:VMT-Adapter}
\paragraph{The Architecture of VMT-Adapter.}
The core of VMT-Adapter is to share the knowledge from multiple tasks to enhance cross-task interaction meanwhile remain specific knowledge for each task with minimal additional trainable parameters. As illustrated in Figure~\ref{fig:adapters}c, VMT-Adapter comprises a shared down-projection layer $W_{down} \in R^{d \times k}$, a nonlinear activation function ReLU, a shared up-projection $W_{up} \in R^{k \times d}$, and independent task-specific knowledge extraction modules. The shared projections facilitate interaction across multiple tasks, resulting in cross-task latent features $F$, as follows:
\begin{equation}
F=\operatorname{ReLU}\left(x_{\ell} \cdot W_{down}\right) \cdot W_{up},
\end{equation}
where $x_{\ell}$ represents the input of $\ell$-th layer. Subsequently, we partition the cross-task features into two components using a gating ratio. The first component represents the task-generic knowledge, while the second component is fed into the knowledge extraction modules. These modules perform scaling and shifting operations via dot product to obtain task-specific representations for each task, as follows:
\begin{equation}
f_{t_i}= \alpha_i  \odot (s \cdot F) + \gamma_i,
\end{equation}
where $s$ is the gating ratio, $\alpha_i \in R^{d}$ and $\gamma_i \in R^{d}$ denote the scale and shift factors for $i$-th task. $\odot$ is the dot product.

\paragraph{How to use VMT-Adapter.} As shown in Figure~\ref{fig:background}, we insert VMT-Adapter in parallel to the MLP in each transformer layer. For multi-task dense scene understanding, the decoder of each task receives multi-scale information from different transformer blocks. Therefore, task-specific representations generated by VMT-Adapter in each layer is directly transmitted to the decoder as a part of the multi-scale feature. The task-generic representations of the VMT-Adapter is added to the encoder and transferred to the decoder after each block. For the decoder of $i$-th task, the received multi-scale features from encoder are expressed as:
\begin{equation}
    F^{ms}_{i} = [(f_g^1 + f_{t_i}^1),\; (f_g^2 + f_{t_i}^2),\; (f_g^3 + f_{t_i}^3),\; (f_g^4 + f_{t_i}^4)],
\end{equation}
where $f_g^j$ and $f_{t_i}^j$ denote task-generic and task-specific representations from the $j$-th block. 

\subsection{A More Lightweight VMT-Adapter}
\label{section:VMT-Adapter-Lite}
In accordance with the VMT-Adapter architecture, the trainable parameters of the task-specific knowledge extraction module are almost negligible compared to the two shared projections. This implies that adjusting $W_{down}$ and $W_{up}$ can further minimize the additional parameters, thereby meeting more stringent efficiency requirements. Therefore, we propose a parameter-sharing method to further decrease the number of trainable parameters. 

We generate $W_{down}$ and $W_{up}$ using a set of shared matrices $\mathbf{A}=\{A^{i}|1\le i \le m\}$, down-specific matrices $\mathbf{B}_{down}=\{B^{i}_{down}|1\le i \le m\}$, and up-specific matrices $\mathbf{B}_{up}=\{B^{i}_{up}|1\le i \le m\}$, by computing the sum of Kronecker products as follows:
\begin{equation}
W_{down} = \sum_{i=1}^{m} A^{i}\otimes B^{i}_{down}\, ; \, W_{up} = \sum_{i=1}^{m} A^{i}\otimes B^{i}_{up},
\end{equation}
where $A^{i} \in R^{m \times m}$, $B^{i}_{down} \in R^{\frac{d}{m} \times \frac{k}{m}}$ and $B^{i}_{up} \in R^{\frac{k}{m} \times \frac{d}{m}}$. This shared strategy reduces the parameters of the VMT-Adapter to $\frac{1}{m}$ of the original, yielding a more light-weight Adapter termed \textit{VMT-Adapter-Lite}.

\subsection{Discussion}
\label{section:insights}
\paragraph{Trainable Parameter.}
We compare the trainable parameters of Multiple Adapter, Shared Adapter, and VMT-Adapter(-Lite). Taking SwinTransformer as an example, in which the dimension is $d$ and the number of layers is $L$. Assuming that Adapter projects features from $d$-dim to $k$-dim, where $k=\frac{d}{\rho}$ and $\rho$ is the down-projection ratio.

Given $T$ tasks, Multiple Adapter inserts $T$ adapters in each transformer layer. Each adapter consists of $2kd$ parameters for the down- and up-projections. Therefore, the total number of trainable parameters for the SwinTransformer model with $L$ layers is $TL\cdot 2kd$. Shared Adapter inserts single adapter in each transformer layer with $L\cdot 2kd$ parameters. VMT-Adapter introduces independent knowledge extraction module on the basis of Shared Adapter. For $T$ tasks, the parameter number of this module is $2Td$. Consequently, the total parameters is $L\cdot 2kd+2Td$, where $T<<k$. VMT-Adapter-Lite shares the trainable parameter $\{A^i\}^m_{i=1}$ of $m^3$. The parameters for down- and up-projections are reduced to $\frac{kd}{m}$. For a SwinTransformer model with L layers, the total number of parameters of $m^3+L\cdot \frac{2kd}{m}+2Td$. Finally, $k$ is replaced by $\frac{d}{\rho}$, and the results are presented in Table \ref{table: complexity}.
\begin{table}
	\centering
	\small
	\scalebox{0.63}{
		\begin{tabular}{ccccc}
			\toprule
			Method & Multiple Adapter & Shared Adapter & VMT-Adapter & VMT-Adapter-Lite  \\ \midrule
			\# Trainable Params. & $\frac{2TL}{\rho}d^2$ & $\frac{2L}{\rho}d^2$ & $\frac{2L}{\rho}d^2+2Td$ & $m^3+\frac{2L}{m\rho}d^2+2Td$ \\ \midrule
			\# T\&I Efficiency. &$O(T)$  &$O(1)$  &$O(1)$  &$O(1)$  \\ %\midrule
			\bottomrule
	   \end{tabular}
    }
	\caption{The trainable parameters and training/inference (T\&I) efficiency comparisons. $T$ is the task number, L is the layer number, m is the dimension of the parameter-sharing matrix, and $\rho$ is the down-projection ratio.}
    \label{table: complexity}
\end{table}
\paragraph{Training \& Inference Efficiency.}
Since the Multiple Adapter establishes separate paths for each task during both training and inference, each sample must pass through the encoder  $T$ times to obtain predictions for $T$ tasks. Consequently, the training and inference efficiency of the Multiple Adapter, including Polyhistor~\cite{liu2022polyhistor} and Hyperformer~\cite{karimi2021parameterefficient}, is $O(T)$.  In contrast, the VMT-Adapter architecture allows only the task-generic representations to pass through the encoder, while the task-specific representations are computed in parallel. This results in an approximately $O(1)$ training and inference efficiency. In summary, our method not only strikes a balance between trainable parameters and performance but also achieves optimal training and inference efficiency.

\paragraph{Gradient Analysis.} For task $t_i$, the loss function is $\mathcal{L}_i(\theta _{sh}, \theta_{sp}^i)$,  where $\theta _{sh}$ are down- and up-projection parameters shared among all tasks and $\theta_{sp}^i$ are task-specific parameters. We denote the gradient of task  $t_i$ with respect to the shared parameters as $\mathbf{g}_{i}=\nabla_{\theta_{\mathrm{sh}}} \mathcal{L}_{i}\left(\theta_{\mathrm{sh}}, \theta_{sp}^i\right)$ . The effect of this change on another task $t_j$ is measured by:
\begin{figure}[t]
   \begin{picture}(0,257)
     \put(0,140){\includegraphics[width=1\linewidth]{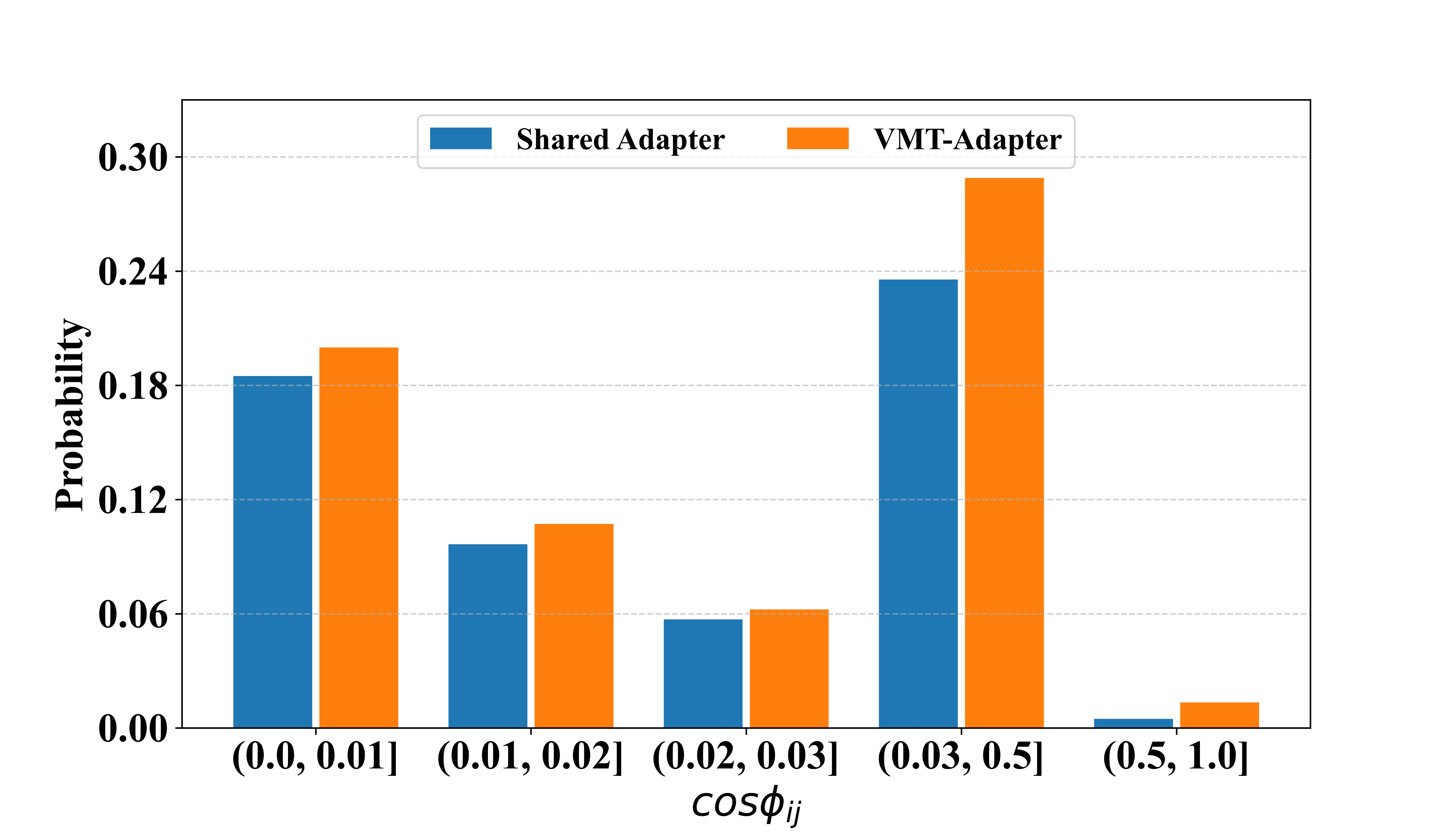}}
     \put(0,5){\includegraphics[width=1\linewidth]{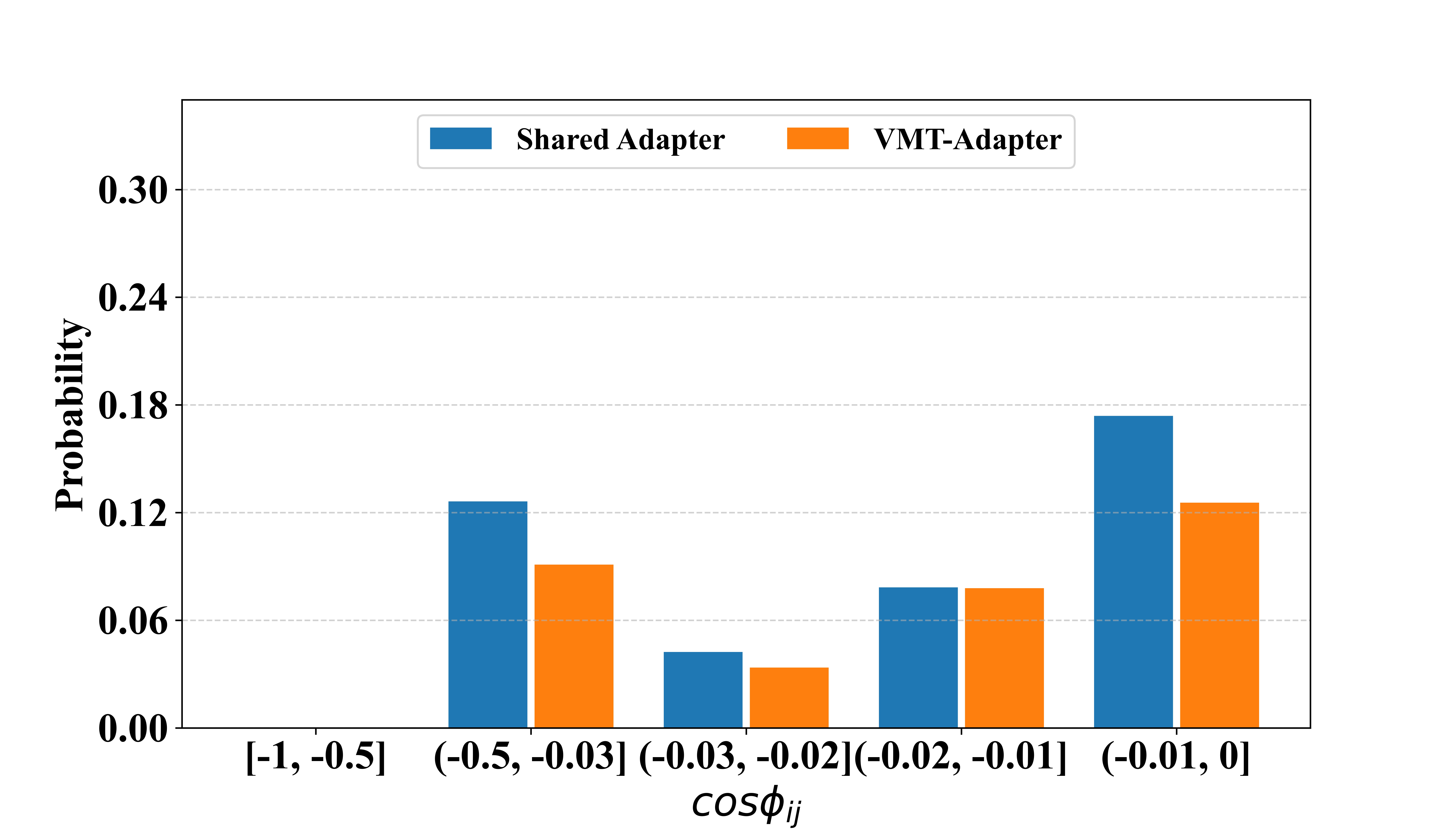}}
     \put(98,130){(a) $cos\phi _{ij} > 0$  }
     \put(98,-3){(b) $cos\phi _{ij} < 0$}
   \end{picture}
    \caption{The gradient conflicts distributions of the Shared Adapter and VMT-Adapter. (a) $cos\phi _{ij} > 0$ represents positive effect and (b) $cos\phi _{ij} < 0$ represents negative effect.}
    \label{fig:gradient}
\end{figure}
\begin{equation}
\begin{aligned}
    \Delta \mathcal{L}_{j}&=\mathcal{L}_{j}\left(\theta_{\mathrm{sh}} - \eta \mathbf{g}_{i}, \theta_{sp}^{j}\right)-\mathcal{L}_{j}\left(\theta_{\mathrm{sh}}, \theta_{sp}^{j}\right)\\
    &=-\eta \mathbf{g}_{i} \cdot \mathbf{g}_{j}+o(\eta), 
\end{aligned}
\end{equation}
where $\eta$ is a sufficiently small step size and the second equality is obtained by first order Taylor approximation. The model update for task $t_i$ is considered to have a positive effect on task $t_j$ when $\mathbf{g}_{i} \cdot \mathbf{g}_{j}>0$. Therefore, the cosine similarity can be used to represent the relationship between tasks, specifically expressed as:
\begin{equation}
    cos\phi _{ij} = \frac{\mathbf{g}_{i} \cdot \mathbf{g}_{j}}{\parallel \mathbf{g}_{i} \parallel \parallel \mathbf{g}_{j} \parallel},
\end{equation}
where $\phi _{ij}$ is the angle between $\mathbf{g}_{i}$ and $\mathbf{g}_{j}$. We then compute conflict angle for the shared parameters between any two tasks (four tasks on PASCAL-Context) in terms of $cos\phi _{ij}$. We then count and draw the distribution of $cos\phi _{ij}$ in all training iterations. As shown in Figure~\ref{fig:gradient}, it can be seen that (a) the probability of $cos\phi _{ij} > 0$ in VMT-Adapter has been improved, increasing the synergy between tasks and (b) the probability of $cos\phi _{ij} < 0$ in VMT-Adapter is reduced, reducing conflict between tasks.

\begin{table*}[t]
\centering
\renewcommand{\arraystretch}{1.3}
\resizebox{0.95\linewidth}{!}{%
\begin{tabular}{ccccccccc}
\toprule
& \multicolumn{1}{c}{Number of Trainable Parameters}         
&  
& \multicolumn{4}{c}{Performance of Each Downstream Task} 
&           
& \multicolumn{1}{c}{Averaged Results} \\
    & Encoder / All (M)
    
    &  & Seg. $\uparrow$   
    & H.Part $\uparrow$  
    & Sal. $\uparrow$  
    & Normals $\downarrow$  &           
&\multicolumn{1}{c}{ $\Delta_{up}$ } \\

\midrule
Single-task Full Fine-tuning & 110.07 / 112.62  &  & 67.21 & 61.93 & 62.35 & 17.97 & & 0.00$\%$ \\

Fine-tuning Decoders & 0.00 / 2.55 &  & 63.14 & 52.37 & 58.39 & 20.89 & &-11.02$\%$ \\

Multi-task Full Fine-tuning  & 27.51 / 30.06 &  & 68.71 & 62.13 & 64.18 & 17.35 & & 2.23$\%$ \\

%%%%%%%%%%%%%%%%%%%%%%%%%%%%%%%%%%%%%%%%%%%%%%%%%
%%%%%%%%%%%%%%%%four times data%%%%%%%%%%%%%%%%%%%%%
%%%%%%%%%%%%%%%%%%%%%%%%%%%%%%%%%%%%%%%%%%%%%%%%%
\midrule

Multiple Bitfit~\cite{zaken2021bitfit} & 0.30 / 2.85 &  & 68.57 & 55.99   & 60.64  & 19.42  & & -4.60$\%$ \\

Multiple Relative bias~\cite{liu2021swin}  & 0.09 / 2.64 &  & 63.51 & 52.35 & 57.74 & 21.07 & & -11.40$\%$ \\

Multiple LoRA~\cite{hu2021lora}                   & 0.32 / 2.87       &  & 70.12       & 57.73        & 61.90      & 18.96         &           & -2.17$\%$                    \\
Multiple Adapter~\cite{he2022towards}    & 8.69 / 11.24               &  & 69.21       & 57.38        & 61.28      & 18.83         &           & -2.71$\%$                    \\
Multiple Low-rank adapter~\cite{sung2022vladapter}                & 0.34  / 2.89               &  & 68.31       & 56.53        & 60.29      & 19.36         &           & -4.54$\%$                    \\ %\midrule

%%%%%%%%%%%%%%%%%%%%%%%%%%%%%%%%%%%%%%%%%%%%%%%%%
%%%%%%%%%%%%%%%%one times data%%%%%%%%%%%%%%%%%%%%%
%%%%%%%%%%%%%%%%%%%%%%%%%%%%%%%%%%%%%%%%%%%%%%%%%
Shared BitFit~\cite{zaken2021bitfit}
& 0.08 / 2.63 & &67.99  &56.23 &60.96 &18.63 & &-3.49$\%$ \\

Shared Relative bias~\cite{liu2021swin}
& 0.03 / 2.58 & &65.55  &54.44 &59.14 &19.52 & &-9.56$\%$ \\

Shared LoRA~\cite{hu2021lora}
& 0.13 / 2.68 & &68.96  &56.71 &61.07 &17.92 & &-1.90$\%$ \\

Shared Adapter~\cite{he2022towards} 
& 2.17 / 4.71 & &70.21  &59.15 &62.29 &19.26 & &-0.64$\%$ \\

%%%%%%%%%%%%%%%%%%%%%%%%%%%%%%%%
Shared Low-rank adapter~\cite{sung2022vladapter}
& 0.10 / 2.62 & &66.84  &55.52 &60.21 &18.51 & &-4.33$\%$ \\

\midrule

Hyperformer~\cite{karimi2021parameterefficient}  & 72.77 / 75.32   &  & \textbf{71.43}     & \textbf{60.73}        & \textbf{65.54}      & 17.77         &           & \textbf{2.64}$\%$                  \\ %\midrule
Polyhistor~\cite{liu2022polyhistor}       & 6.41 / 8.96     &  & 70.87       & 59.54        & 65.47      &17.47        &           & 2.34$\%$                    \\

Polyhistor-lite~\cite{liu2022polyhistor}   & 0.41 / 2.96               &  &70.24       & 59.12        & 64.75      & \textbf{17.40 }       & \textbf{} & 1.74$\%$                   \\

%%%%%%%%%%%%%%%%%%%%%%%%%%%%%%%%%%%%%%%%%%%%%%%%%
%    Ours
%%%%%%%%%%%%%%%%%%%%%%%%%%%%%%%%%%%%%%%%%%%%%%%%%
\midrule
\textbf{VMT-Adapter}               
& 1.13 / 3.68              
&  & \textbf{71.60}  & \textbf{60.67}        & 64.02      & \textbf{16.41} 
&  &\textbf{3.96}$\%$ \\

\textbf{VMT-Adapter-Lite}               
& 0.40  / 2.95    
&  & 70.03  & 59.51 & 62.45 & 17.09
&  & 1.34$\%$ \\

\bottomrule
\end{tabular}
}
\caption{
Experimental results on Multi-Task Transfer Learning. We use SwinTransformer-Tiny as the encoder. $\Delta_{up}$ represents relative improvement against the Single-task Full Fine-tuning. Results with the symbol $\uparrow / \downarrow$ indicate higher/lower is better.}
\label{table:pascal-swint}
\end{table*}

\section{Experiment}
\label{sec:exp}
\subsection{Experimental Settings}
\paragraph{Datasets and Downstream Tasks.} To evaluate our proposed approach for multi-task dense scene understanding, we follow the prior works~\cite{Vandenhende2021pami,liu2022polyhistor} and conduct experiments on the PASCAL-Context~\cite{vandenhende2020mti} dataset. PASCAL-Context comprises 4,998 and 5,105 images in the training and testing splits, respectively. We evaluate on four dense prediction tasks, including 21-class semantic segmentation, 7-class human part segmentation, surface normals estimation, and saliency detection. We use the mean Intersection-over-Union~(mIoU) metric to evaluate the semantic segmentation, human part segmentation and saliency detection tasks, while the mean error~(mErr) metric is used for the surface normals estimation task.

\paragraph{Model Architecture.} For the encoder, we adopt the SwinTransformer architecture and initialize it with the parameters pre-trained on ImageNet. For the decoder, we use the All-MLP decoder of Segformer~\cite{xie2021segformer} and set up a decode structure for each task. This structure consists of linear layers and bilinear upsampling layers, which enable efficient performance of dense scene understanding tasks. To ensure a fair comparison, we use the same encoder and decoder for all methods, and our experimental settings are consistent with the previous work Polyhistor\cite{liu2022polyhistor}.

\paragraph{Implementation Details.} We conduct all experiments using the PyTorch toolkit on 4 NVIDIA V100 GPUs. For fair comparison, the hyper-parameters of all methods are the same. Specifically, we use batch size 12 and train for 60 epochs for each task. We employ the Adam optimizer with a learning rate $1e^{-4}$ and a weight decay $1e^{-4}$, and the learning rate is linearly decreased with respect to the iteration. 

\subsection{Baselines}
\textbf{Single-task Full Fine-tuning} uses independent pre-trained encoder and decoder for each task, and updates the entire model. \textbf{Fine-tuning Decoders} updates the decoder parameters. \textbf{Multi-task Full Fine-tuning} uses a shared pre-trained encoder and separate decoder, then updates the entire model. 

For single-task parameter-efficient transfer learning methods (\textit{i.e.}, BitFit, Relative bias, LoRA, Adapter, and Low-rank Adapter), Polyhistor proposed a \textbf{Multiple} benchmark, which involves using a separate adaptation method for each task. In contrast, we extend a \textbf{Shared} benchmark, which involves sharing the adaptation method across multiple tasks.

For multi-task settings with multiple adapters, \textbf{Hyperformer} applies a hyper-network to produce weights for the adapters. \textbf{Polyhistor} further uses a scaling module to adapt the shared weights to hierarchical vision transformers.

\begin{table*}[h]
\centering
\renewcommand{\arraystretch}{1.3}
\resizebox{\linewidth}{!}{%
\begin{tabular}{ccccccccccc}
\toprule
&   Size of   & \multicolumn{1}{c}{Number of Trainable Parameters} 
&        
& \multicolumn{4}{c}{Performance of Each Downstream Task} 
&      
& \multicolumn{1}{c}{Averaged Results} \\
& down-projection ratio $\rho$                
& Encoder / All (M)                 
&  
& Seg. $\uparrow$   
& H.Seg. $\uparrow$  
& Sal. $\uparrow$  
& Normals $\downarrow$  
&           
& $\Delta_{up}$ \\ \midrule
Single-task Full Fine-tuning    &        -      & 110.07 / 112.62           &  & 67.21               & 61.93        & 62.35      & 17.97         &           & 0.00$\%$      \\           
Fine-tuning Decoders &        -          & 0.00 / 2.55        &  & 63.14       & 52.37        & 58.39      & 20.89         &           & -11.02$\%$  \\ \midrule
\textbf{VMT-Adapter}   & $\rho=1$     & 4.36 / 6.91               &  & 71.26       & 61.01        & 63.76      & 16.12         & \textbf{} & \textbf{4.27}$\%$  \\
\textbf{VMT-Adapter}   & $\rho=2$     & 2.21 / 4.56               &  & 70.96       & 60.82        & 64.01      & 16.39         & \textbf{} & \textbf{3.81}$\%$                  \\
\textbf{VMT-Adapter}   & $\rho=4$     & 1.13 / 3.68              &  & 71.60       & 60.67        & 64.02      & 16.41         & \textbf{} & \textbf{3.96}$\%$                  \\
\textbf{VMT-Adapter}   & $\rho=8$     & 0.59 / 3.14              &  & 71.34       & 59.95        & 63.48      & 16.82         & \textbf{} & \textbf{3.02}$\%$                  \\

\midrule
\textbf{VMT-Adapter-Lite}   & $m=3$     & 0.40  / 2.95    
&  & 70.03  & 59.51 & 62.45 & 17.09
&  & \textbf{1.34} $\%$ \\
\textbf{VMT-Adapter-Lite}   & $m=6$     & 0.22 / 2.77               &  & 69.85       & 59.38        & 61.95      & 17.53  &  & \textbf{0.61}$\%$                  \\
\textbf{VMT-Adapter-Lite}   & $m=12$     & 0.15 / 2.70              &  & 69.33       & 59.15        & 61.88      & 17.55         & \textbf{} & \textbf{0.06}$\%$                  \\
\bottomrule
\end{tabular}
}
\caption{
Ablation study on the ratio of down-projection of VMT-Adapter and parameter-sharing matrix dimension $m$ of VMT-Adapter-Lite. We vary ratio $\rho$ from 1 to 8 and $m$ from 3 to 12 ($\rho=4$).}
\label{table:different ratio}
\end{table*}

\begin{figure*}[t]
   \begin{picture}(0,125)
     \put(20,-4){\includegraphics[width=0.45\linewidth]{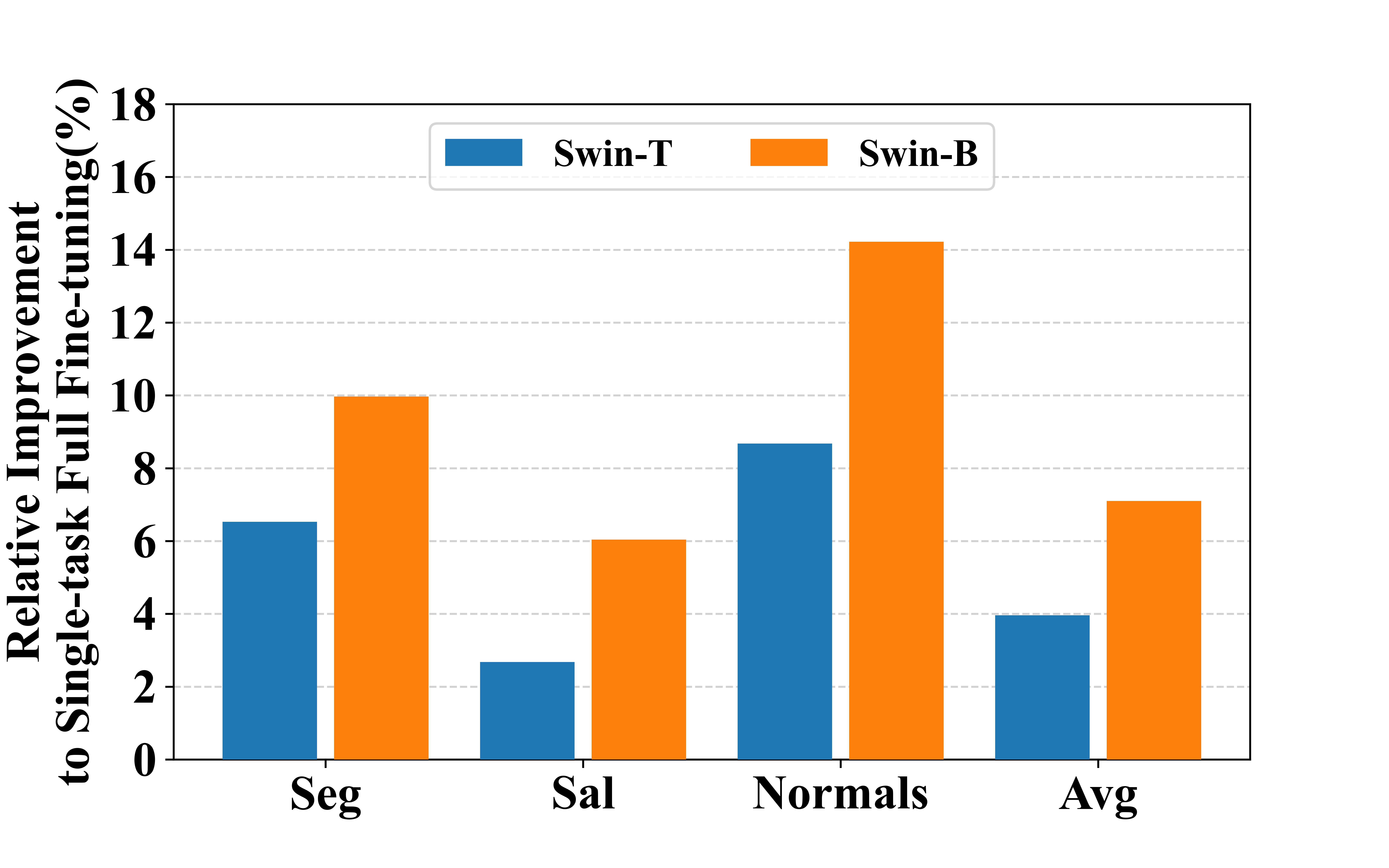}}
     \put(260,-4){\includegraphics[width=0.45\linewidth]{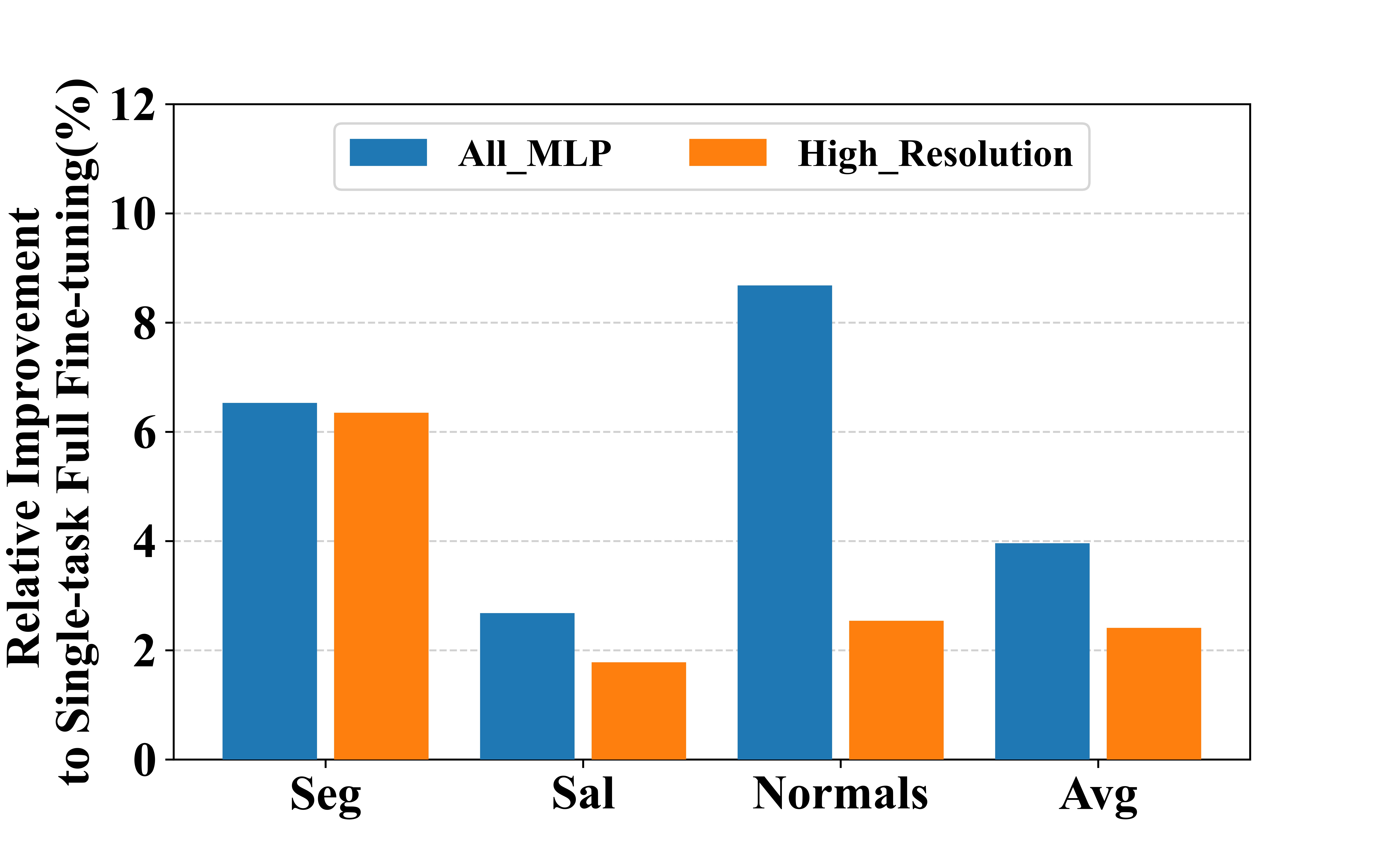}}

     \put(87,-6){(a) Different Encoder}
     \put(330,-6){(b) Different Decoder}

   \end{picture}
    \caption{Ablation study on (a) different encoder and (b) different deocder. All results are produced by VMT-Adapter.}
    \label{fig:ablation_study}
\end{figure*}
\subsection{Performance Comparisons} 
We evaluate all methods on four dense scene understanding tasks. In addition to showing the performance metrics of each task, we also evaluate the relative improvement against Single-Task Full Fine-tuning. At last, the number of trainable parameters is also reported.

As presented in Table~\ref{table:pascal-swint} and Figure~\ref{fig:pipeline}, \textit{VMT-Adapter} outperforms all other methods with an average improvement of +3.96\% on four downstream tasks against Single-task Full Fine-tuning, while leveraging only 1.13M trainable parameters. VMT-Adapter-Lite reduces the trainable parameters to 0.40M and still achieves a +1.34\% improvement. Hyperformer performs the best besides our method with +2.64\% improvement. However, Hyperformer introduces considerable additional parameters, i.e., 72.77M that even exceeds the encoder with 27.51M parameters, violating the principle of parameter-efficient transfer learning. Polyhistor performs well with a +2.34\% average improvement on four downstream tasks, leveraging 6.41M trainable parameters (about 6 times as much as ours). Polyhistor-Lite further reduces the number of trainable parameters while ensuring performance. However, both methods use task-independent adapters, leading to $O(T)$ training and inference efficiency. 

For single-task adaptation methods (\textit{e.g.}, BitFit, LoRA, Adapter, and Low-rank Adapter), despite some methods having fewer trainable parameters, they cannot relieve the performance drop, resulting in a significant gap from -0.64\% to -11.40\% against Single-Task Full Fine-tuning. Additionally, it is worth noting that the Shared Adapter performs better than the Multiple Adapter, with even competitive results on par with Single-Task Full Fine-tuning, indicating that collaboration between tasks can improve the performance of each task.

\subsection{Ablation Studies}
\paragraph{Different Pre-trained Encoders.} To verify the generalization of our method, we conduct experiments on a larger backbone, SwinTransformer-Base, pre-trained on ImageNet. The results are shown in Figure~\ref{fig:ablation_study}a. According to the results, the performance of four tasks is significantly improved compared with the Single-task Full Fine-tuning (+3.96\% average improvement of SwinTransformer-Tiny and +7.10\% of SwinTransformer-Base). We found that \textit{VMT-Adapter} works better on larger backbone, showing that our method is applicable to various backbones for multi-task dense scene understanding.

\paragraph{Different Decoders.} The decoder is an essential part for dense scene understanding tasks. In order to prove that our method is not benefiting from a specific decoder structure, we further conduct experiments on the high-resolution decoder of HRNet-V2~\cite{sun2019high}, which aggregates the representations at different resolutions. For a fair comparison, we use SwinTransformer-Tiny as the encoder structure. As shown in Figure~\ref{fig:ablation_study}b, \textit{VMT-Adapter} achieves +3.96\% and +2.41\% relative improvements using All-MLP and High-Resolution decoders, respectively. Therefore, \textit{VMT-Adapter} is flexible and can be adapted to various decoders. 

\paragraph{Down-Projection Ratio.} The down-projection ratio of our \textit{VMT-Adapter} is a crucial hyper-parameter, thus we try different ratios $\rho=\frac{d}{k}$. As presented in Table \ref{table:different ratio}, we vary the ratio $\rho$ from 1 to 8 based on SwinTransformer-Tiny. The experimental results show that different $\rho$ improves performance from 3.02\% to 4.27\%, using only 0.5\% to 3.9\% trainable parameters of Single-task Full Fine-tuning.

\paragraph{Parameter-Sharing Matrix Dimension.} We also examine how our proposed \textit{VMT-Adapter-Lite} performs when different parameter-sharing matrix dimensions are used. From Table~\ref{table:different ratio}, we observe that with the increase of $m$, the number of parameters gradually decreases, but the performance of multiple tasks also decreases. Therefore, in practical applications, we recommend setting m to 3, which can achieve a trade-off between performance and trainable parameters.

\section{Conclusion}
\label{sec:conclusion}
In this work, we propose \textit{VMT-Adapter(-Lite)} for vision multi-task parameter-efficient transfer learning, which can simultaneously learn task-generic and task-specific representations from different transformer blocks. Compared with Single-task Full Fine-tuning and other parameter-efficient transfer learning methods, \textit{VMT-Adapter(-Lite)} achieves favorable results while using a limited number of tunable parameters. In addition, \textit{VMT-Adapter(-Lite)} is also optimal regarding training/inference costs. The potential limitation of \textit{VMT-Adapter} is that the task-specific module parameters will be higher than the shared parameters when the number of tasks reaches thousands. 

\bibliography{aaai24}

\end{document}